\documentclass{article}

\usepackage{booktabs}
\usepackage{PRIMEarxiv}
\usepackage{listings}
\usepackage{courier}
\usepackage{mathpazo} 
\usepackage[utf8]{inputenc} 
\usepackage[T1]{fontenc}    
\usepackage{hyperref}       
\usepackage{url}            
\usepackage{booktabs}       
\usepackage{amsfonts}       
\usepackage{nicefrac}       
\usepackage{microtype}      
\usepackage{lipsum}
\usepackage{url}
\usepackage{fancyhdr}       
\usepackage{graphicx}       
\usepackage{tcolorbox}

\usepackage[backend=biber, style=numeric-comp, sorting=none, doi=true, eprint=true, url=true]{biblatex}
\graphicspath{{media/}}     

\newcommand{\authorblock}[3]{
  \noindent
  \textbf{#1, #2, and #3}\\
  Department of Physics, Durham University, United Kingdom\\
  *Email: \href{mailto:will.yeadon@durham.ac.uk}{will.yeadon@durham.ac.uk}
}

\makeatletter 
\renewcommand{\@maketitle}{
  \vbox{%
    \hsize\textwidth
    \linewidth\hsize
    \vskip 0.1in
    \@toptitlebar
    \centering
    {\LARGE \@title\par}
    \@bottomtitlebar
    \vskip 0.2in 
    {\large \authorblock{Will Yeadon}{Craig P. Testrow}{Alex Peach}} 
    \vskip 0.3in 
  }
}
\makeatother
\newcommand\goldilocks[1]{
\fontsize{21}{30}\selectfont{#1}} 

\title{\goldilocks{A comparison of Human, GPT-3.5, and GPT-4 Performance in a University-Level Coding Course}}
\author{Will Yeadon, Alex Peach, Craig P. Testrow}
\begin{document}
\maketitle

\begin{abstract}
This study evaluates the performance of ChatGPT variants, GPT-3.5 and GPT-4, both with and without prompt engineering, against solely student work and a mixed category containing both student and GPT-4 contributions in university-level physics coding assignments using the Python language. Comparing 50 student submissions to 50 AI-generated submissions across different categories, and marked blindly by three independent markers, we amassed $n = 300$ data points. Students averaged 91.9\% (SE:0.4), surpassing the highest performing AI submission category, GPT-4 with prompt engineering, which scored 81.1\% (SE:0.8) - a statistically significant difference (p = $2.482 \times 10^{-10}$). Prompt engineering significantly improved scores for both GPT-4 (p = $1.661 \times 10^{-4}$) and GPT-3.5 (p = $4.967 \times 10^{-9}$). Additionally, the blinded markers were tasked with guessing the authorship of the submissions on a four-point Likert scale from `Definitely AI' to `Definitely Human'. They accurately identified the authorship, with 92.1\% of the work categorized as 'Definitely Human' being human-authored. Simplifying this to a binary `AI' or `Human' categorization resulted in an average accuracy rate of 85.3\%. These findings suggest that while AI-generated work closely approaches the quality of university students' work, it often remains detectable by human evaluators.
\end{abstract}


\section{Introduction}
Coding courses are now ubiquitous in university curricula globally, signifying the increasing recognition of programming as a critical skill within the digital economy. The emergence of advanced Large Language Models (LLMs), such as Codex that powers GitHub Copilot \cite{codex}, prompts a reevaluation of coding assessments' efficacy and integrity within educational settings. Whilst the coding abilities of LLMs have been well-explored through benchmarks like MBPP and MathQA-Python \cite{programBench}, our investigation focuses on AI's potential effects on university coding courses. Rather than evaluating AI's capacity to solve coding puzzles, as seen in work by Tian et al. \cite{tian2023chatgpt}, this study looks at AI's potential impact on a practical coding curriculum. Specifically, we consider a 10-week coding course within a physics degree at Durham University, where students engage in lectures and complete assignments on an nb-grader powered Jupyter notebook server.

Physics degrees offer a broad spectrum of assessments, ranging from lab-based experiments and presentations to written exams, essays, and coding assignments. This variety distinguishes physics from disciplines that rely more heavily on particular types of assessments, such as written exams within mathematics, a greater focus on lab work in chemistry, a more essay-centric approach in English literature, or computer science where looking at student's coding abilities is a key form of assessment. This multifaceted approach within a single subject provides a good opportunity to assess AI's impact on various forms of academic assessment. Arguably, the tangible, hands-on nature of lab work and presentations largely precludes AI’s influence. While AI's role in written physics exams has yet to match students \cite{Yeadon_2024}, the integrity of essay-based assessments is being increasingly challenged by AI's ability to produce work that is both indistinguishable from, and comparable in quality to, that of humans \cite{yeadon2024evaluating}. Further, a critical aspect of a physics degree involves how students grasp, interpret, and apply core physics principles. Recent studies probing LLMs’ comprehension of physics indicate their performance is improving significantly, with results increasingly approximating human levels; however, peculiar errors persist \cite{westy, Kortemeyer, BG-kinematics, BG2}. This nuanced progress highlights the necessity to continuously reassess the role and effectiveness of coding assignments as AI technologies advance. By examining physics coding courses, this study seeks to determine the ongoing validity, utility and integrity of such assignments in accurately assessing student performance in an era of rapid technological development. To ensure transparency and enable replication, the code used in this research is openly available on GitHub\footnote{https://github.com/WillYeadon/AI-Exam-Completion}.

\section{Methodology}
\label{sec:method}
\subsection{Overview}
This study aims to assess the effectiveness of contemporary Large Language Models (LLMs) in performing coding tasks typically assigned to university students, using a blinded marking approach to evaluate code written by both students and AI. In the context of physics, coding is primarily utilized for simulations and data analysis, including plotting. A critical skill is therefore the creation of clear, well-labeled plots that elucidate the underlying physics of a scenario. Consequently, this physics coding assessments place special emphasis on the quality of the produced plots and the performance of the code used to create them, particularly in terms of runtime. This approach contrasts with Computer Science scenarios, where students are expected to explain and justify their coding choices, with readability and maintainability being key evaluation criteria. To determine whether ChatGPT is an effective coding tool for physics education, 14 plots, documented in a 16-page report, were evaluated against a specific marking scheme for both AI and student-authored submissions. After blinded marking, the evaluators assigned the probably authorship on a 4-point Likert scale from `Definitely AI' to `Definitely Human'.

\subsection{Coding Assignment}
The coding assignments using in this study come from the `Laboratory Skills and Electronics' module at Durham University, UK, designed for physics and natural sciences students in their second and third years. This module encompasses essential laboratory practices, electronics, and a coding component. The Python coding segment spans 10 weeks, featuring weekly lectures and a total of 8 weekly assignments, excluding the first and last week. Topics covered include finite difference methods, numerical integration, solving first and second order differential equations, Monte Carlo methods, and random walks. We solicited submissions from the 2023/24 student cohort, and from 103 consenting participants, 55 were randomly selected to represent the body of student work. A limitation here is that there is a chance the submitted student work may be in part itself be generative AI. This was explicitly against the course rules and warnings against the use of generative AI were made multiple times during the course. However, we can't rule this out.

The 8 assignments are structured as a series of online Jupyter notebooks with short tasks that account for 30-60\% of the total marks based on the completion of functions (graded automatically using assert statements) and the remainder assessed through manually graded plotting tasks. The course involves producing a total of 14 plots across the eight assignments: one plot for each of the first three assignments, two plots for each of the following four assignments, and three plots for the final assignment. Each assignment is worth one-eighth of the total coding mark and are scored out of 20. The plots themselves have varying marks assigned to them so for the purpose of this study each plot was scored out a five giving a max score of 70.

\subsection{Generating the AI Code}
While students complete the assignments on an online nbgrader-powered Jupyter notebook server, enabling easy extraction of their plots as images, simulating AI completion of these notebooks requires converting them into a text format for input into an LLM. However, without explicit instructions such as `\emph{Please complete this assignment}', there is no guarantee that an LLM will interact with the submitted text appropriately. Within many of the assignment notebooks students are often given a small amount of starter code, without specific guidance this can lead to the AI redefining variables or even data that has already been provided. Additionally, as previously mentioned, the assignments allocate 40-60\% of the marks to assert statements, including hidden cells that are irrelevant when evaluating plots. Moreover, although there are 8 assignments yielding 14 plots, if the LLM produces code that throws errors, it should not automatically result in a zero score for the other plots within an assignment. These issues pose significant challenges for ensuring a fair comparison between AI-generated and student-generated submissions. Therefore, the assignment notebooks were pre-processed when converted into input for the LLM to align with AI processing capabilities. However, this act somewhat complicates a direct comparison as it means assisting the LLM, thus possibly altering the 'human versus AI' dynamic into 'human versus AI with human assistance.' Given the manner in which pre-processing is conducted has been found to influence the performance of GPT-4 \cite{fengAI}, we implemented only those changes that were absolutely necessary to get the AI to consistently complete the assignments. These changes are detailed in Table \ref{changes}.

\begin{table}[htb]
\centering
\caption{Summary of the pre-processing changes to the raw .ipynb files to transfer them into Python scripts that could be completed by LLMs to compare to student work.}
\label{changes}
\begin{tabular}{clp{12cm}} 
\toprule
\textbf{Change} & & \textbf{Description} \\
\midrule
1 & & All graded cells tested via assert statements up to the plotting tasks were completed with correct answers, these cells represent 30-60\% of the marks per notebook. \\
\addlinespace
2 & & At the beginning of the script, the text `\emph{\# Please read this script and then complete the plot described by `::: Task :::' writing your code where the script indicates `HERE HERE HERE'}’ was inserted. The `::: Task :::' and `HERE HERE HERE' were strategically placed to direct the LLM to the task and the specific location for code input, respectively. This approach was adopted to ensure that the AI consistently responds to the script rather than ignoring it or asking for more information. \\
\addlinespace
3 & & Cells containing assert statements for autograding were excluded from API submissions since the necessary code to pass them was already provided. Moreover, some of these cells were hidden from students anyway. This was done as the details of the assert statements are only pertinent for behind-the-scenes operation and are not relevant for the students or the AI. \\
\addlinespace
4 & & Some cells contained tables that the API couldn't parse in their original form, so these tables were converted into Python dictionaries. This alteration was solely for formatting purposes to facilitate parsing and did not involve any change in the data or addition of new information. \\
\addlinespace
5 & & Course-related content present in the notebooks, such as links to the course website, copied material, or submission guidelines for the internal server, was removed. Additionally, instructions like ‘click on the ``+'' button to create new cells’ were omitted for being considered unnecessary and potentially confusing to the AI. These edits were made to focus the content more directly on the study's objectives. \\
\addlinespace
6 & & In instances where multiple plots were required in a single workbook, AI-generated solutions were inserted at the appropriate places. Surrounding text was minimally altered for clarity, changing phrases like `Create a plot’ to ‘Below is a simulation’ to better reflect the content and context of the workbook. \\
\bottomrule
\end{tabular}
\end{table}

After the application of any form of pre-processing, it essentially opens ``Pandora's box'' as with enough pre-processing exceptional performance can be achieve despite it not being reflective of actual AI capabilities. Nonetheless, despite the pre-processing steps detailed in Table \ref{changes}, the inputs fed into the LLM still contained a fair amount of clutter. Instructions intended to guide humans through the notebook, such as `\emph{Now implement a function}', could potentially confuse an LLM. The significant enhancement of performance through prompt engineering is well-documented \cite{openai_best}. Consequently, we prepared a second set of inputs in addition to the minimal pre-processing outlined in Table \ref{changes}. These inputs incorporate prompt engineering focused enhancements, as detailed in Table \ref{mods}. Finally, taking the process a step further by merging human and AI work into single submissions, we created the following five categories of AI-generated work:

\begin{itemize}
    \item \textbf{GPT-3.5 raw}: This entry includes the assignment text with the minimal adjustments specified in Table \ref{changes}, submitted directly to OpenAI's API using the gpt-3.5-turbo model for processing.
    \item \textbf{GPT-3.5 with prompt engineering}: Here, the assignment text is modified following the guidelines in Table \ref{mods} to optimize interaction with the GPT-3.5 model, aiming for improved responses.
    \item \textbf{GPT-4 raw}: Adopts the same approach as the `GPT-3.5 Raw' but employs the more advanced gpt-4-1106-preview model.
    \item \textbf{GPT-4 with prompt engineering}: Mirrors the process used for the `GPT-3.5 with prompt engineering', but uses the gpt-4-1106-preview model.
    \item \textbf{Mixed}: This category amalgamates contributions from both students and GPT-4 enhanced with prompt engineering. Out of 10 mixed submissions, 14 plots were selected at random from five student entries using Python's random.choice method to fill half the slots, while the remaining plots were generated by the GPT-4 with prompt engineering.
    \item \textbf{Student}: This group contained 50 student submissions from the randomly selected 55 of the 2023/24 cohort.
\end{itemize}

For each of these categories, we generated 10 submissions (each a PDF document containing 14 plots), resulting in a total of 50 AI submissions. Along with the 50 student submissions previously mentioned, this yielded a total of 100 submissions. These were blindly evaluated by three independent markers, providing $n = 300$ data points. The AI submissions were crafted by sending the input text - either in its raw form or after prompt engineering - to OpenAI's API. The response was then appended to the input text, executed as a Python script, and any resulting plots were extracted as images. At times, the AI's responses included written text or non-Python code, leading to errors where no image could be extracted. To achieve 10 submissions per category, the scripts were run in a while loop until the target number was met. Furthermore, in some cases, actual blank figures were generated. These were excluded as they provided no valuable information for this study. The code used in this research is available on GitHub\footnote{https://github.com/WillYeadon/AI-Exam-Completion}.

\begin{table}[htb]
\centering
\caption{Prompt engineering steps used for the \textbf{GPT-3.5 with prompt engineering}, \textbf{GPT-4 with prompt engineering}, and \textbf{Mixed} categories.}
\label{mods}
\begin{tabular}{clp{12cm}} 
\toprule
\textbf{Change} & & \textbf{Description} \\
\midrule
1 & & Function definitions within the notebooks were rewritten for clarity. For example, textual descriptions within the notebooks such as `\emph{Define the function `f`, such that $\textrm{f}(x) \equiv x^{2}\cos(2x)$. This is the function that we will be integrating.}' were simplified to `\emph{definition of the function $f(x) = x^2 * cos(2x)$}'. \\
\addlinespace
2 & & All non-task-related information was removed to focus solely on the assignment requirements and to avoid potential confusion or distraction. \\
\addlinespace
3 & & An enhanced preamble was added to clearly outline the task and instructions for completion. This included explaining the task, providing objectives, and offering suggestions for successfully completing the assignment. Additionally, explicit locations for code insertion were marked with `\emph{HERE HERE HERE}', guiding both the AI to the expected areas of input. \\
\addlinespace
4 & & Post-task details were elaborated to further guide the completion process. This involved clarifying the task's aim, specifying objectives such as plotting differences between analytical and numerical derivatives, and offering suggestions to enhance the clarity and effectiveness of the plots. \\
\bottomrule
\end{tabular}
\end{table}

\section{Results}
\subsection{Score comparison}
\begin{figure}[!htp]
\centering
\includegraphics[width=15cm]{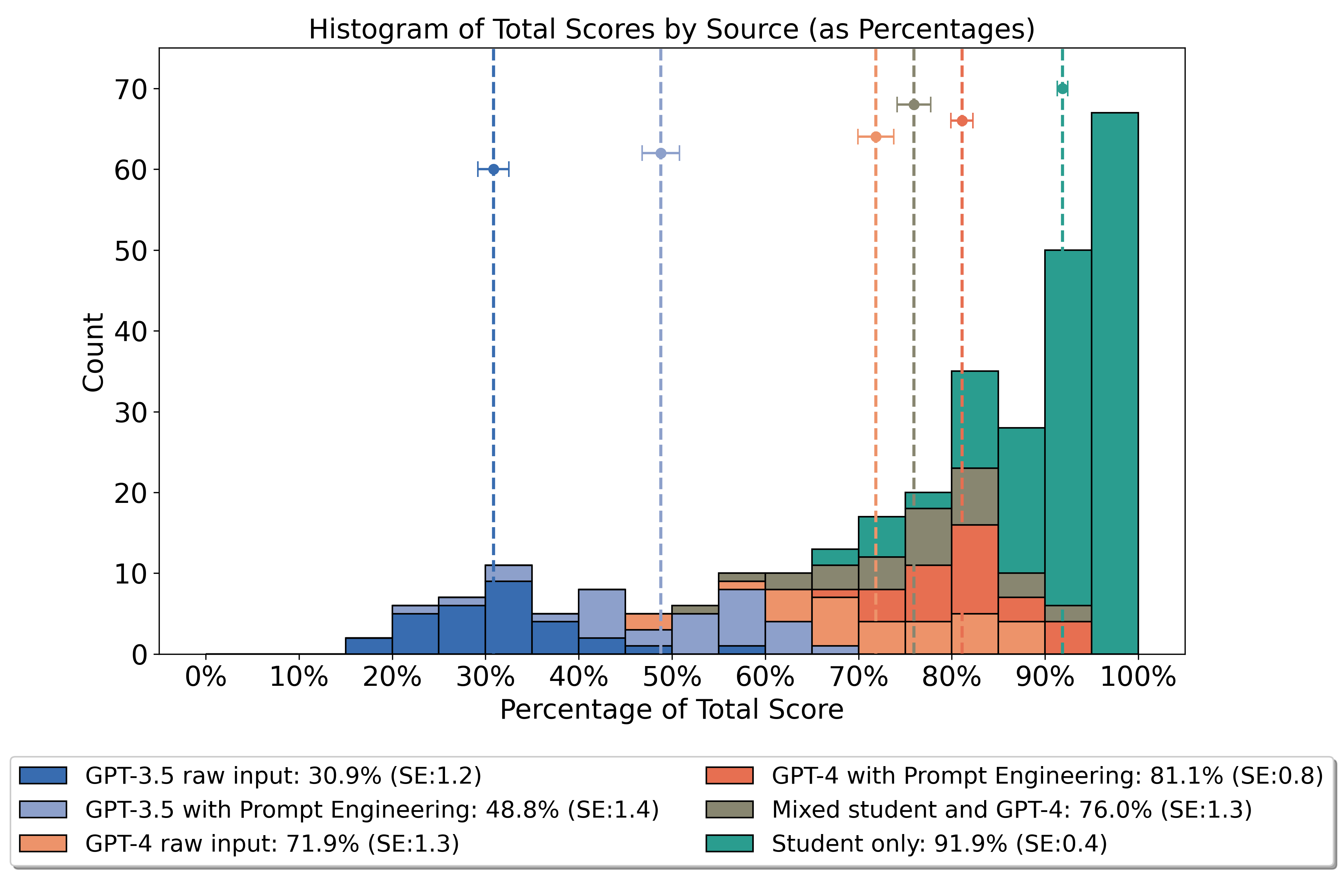}
\caption{Percent scores for each of the six categories of submission. Student submissions score the best thou they are closely followed by GPT-4 with prompt engineering and the Mixed student and AI work. GPT-3.5 performs strictly worse than GPT-4.}
\label{fig-scores}
\end{figure}

A combined dataset of the scores from the three markers for all submissions, evaluated blindly, is shown is shown in Figure \ref{fig-scores}. Here we see students achieved an average of 91.1\% which is in line with the typical average for the actual coding component of `Laboratory Skills and Electronics' at Durham University. In comparison the best performing AI category, GPT-4 with prompt engineering, scores 81.1\%. A t-test between these groups produced a t-statistic of -8.193, with a p-value of $2.482 \times 10^{-10}$. This result shows that although GPT-4 exhibits remarkable capabilities, when it comes to physics coding assignments, it still often isn't as proficient as university students.

Examining the impact of prompt engineering reveals statistically significant improvements: GPT-4's scores increased from 71.9\% (SE:1.3) to 81.1\% (SE:0.8) with a p-value of $1.661 \times 10^{-4}$ from a t-test, and GPT-3.5's scores improved from 30.9\% (SE:1.2) to 48.8\% (SE:1.4) with a t-test giving a p-value of $4.967 \times 10^{-9}$. Thus, as expected, there is are clear and significant benefits to prompt engineering. Interestingly, the mixed submissions, comprising both student and GPT-4 work, scored lower (76.0\% with SE: 1.3) than GPT-4 submissions alone. This may be attributed to variability in the quality of student work sampled, given that the mixed group included plots from five student submissions compared to the 50 in the student-only group.

\subsection{Author identification}
\label{sec-id}
\begin{figure}[!htp]
\centering
\includegraphics[width=15cm]{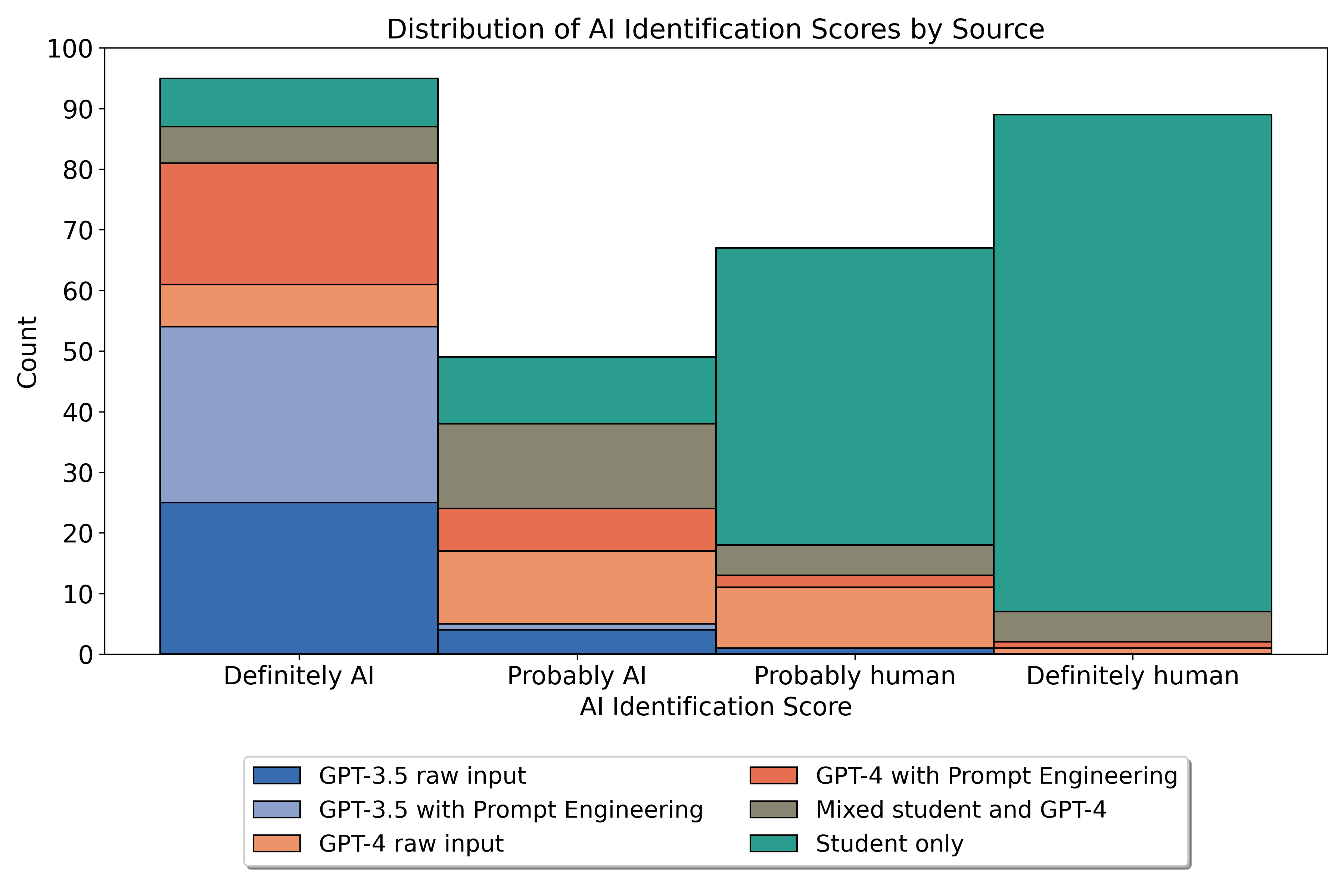}
\caption{Histogram showing the markers’ assigned authorship versus actual authorship of the 300 assessed submissions. The amount of actual human-authored code in ’Definitely human’ is 92.1\% , then 73.1\% in ’Probably human’ followed by 22.4\% in ’Probably AI’ and 8.4\% in ’Definitely AI’.}
\label{fig-id}
\end{figure}

After reviewing each submission, the evaluators assigned authorship scores on a Likert scale, the findings of which are depicted in Figure \ref{fig-id}. This demonstrates that genuine student submissions are more often recognized as student-authored. Converting the Likert scale to a numerical range - assigning `Definitely AI' a value of 0 and `Definitely human' a value of 3 - we arrive at the average scores: 0.033 for GPT-3.5 with raw input, 0.200 for GPT-3.5 with prompt engineering, 0.467 for GPT-4 with raw input, 1.167 for GPT-4 with prompt engineering, 1.300 for the Mixed category (including both human and AI work), and 2.367 for solely human-created work. Therefore all work with an AI-authored component to it has an average categorization closest to either `Definitely AI' (0) or `Probably AI' (1). Furthermore, by designating any submission not solely by students as 'AI-authored' to a degree, and comparing it to 'human-authored' submissions, we applied a Cochran-Armitage test. This test resulted in a p-value of 0.025 and a positive trend of 0.302, statistically verifying that as we move from 'Definitely AI' to 'Definitely Human' on the scale, the proportion of human-authored submissions increases.

These findings indicate that AI-generated content can be identified with a reasonable degree of accuracy. When categorizing the Likert scale responses into a binary system of 'AI' or 'Human,' the reviewers managed an average accuracy rate of 85.3\%, with scores of 89\% for Marker \#1, 77\% for Marker \#2, and 90\% for Marker \#3. It was further observed that content produced by the more advanced GPT-4 model is closer to that created by humans, especially when enhanced through prompt engineering techniques. This observation is contextualized by Figure \ref{fig-scores}, which shows that content generated by humans is generally of higher quality than that produced by any form of AI, implying that work of superior quality is more frequently categorized as human.

\section{Discussion}
\subsection{Overview and recommendations}
The findings of this study indicate that, in the short term, the most sophisticated AI models might not rival human expertise in university-level physics coding assignments. However, our research highlights a distinct improvement of GPT-4 over GPT-3.5, a result also found across diverse disciplines from medicine \cite{medical} to university entrance exams \cite{nunes2023evaluating}. This pattern showcases the steady advancement of AI capabilities, suggesting an evolving landscape where AI's potential to match or surpass human performance becomes increasingly plausible. Given this potential trajectory, it is important for educators to reassess the role of coding assignments and, more broadly, the objectives of their educational strategies. Coding, inherently a practical skill, involves regular consultation of documentation and the reuse of existing code. Therefore, the integration of AI into educational practices, akin to a pair programming setup exemplified by tools like GitHub Copilot, should not be viewed negatively \cite{pairProgramming, github-copilot}. This approach is an exciting opportunity for innovation in physics education. It prompts the exploration of whether AI could surpass traditional teaching methods as a more effective tutor for coding \cite{pairProgrammingGroupWork}.

The lower performance in \textbf{raw input} categories indicates that, from an academic integrity perspective, students are likely to achieve better results by completing the `Laboratory Skills and Electronics' assignments themselves. The extensive prompt engineering for the other AI categories, detailed in Table \ref{mods}, necessitates a level of engagement with the course materials that there is ambiguity regarding its severity from an academic integrity standpoint. In fact, the aforementioned limitation that we could not guarantee the student category was 100\% human written, given it is from 2023/24, may not be a cause of too much worry in this vein as the student work clearly has a different - and higher scoring - distribution in Figure \ref{fig-scores} than both \textbf{Mixed student and GPT-4} and \textbf{GPT-4 with Prompt Engineering}.

While whether a particular amount of `AI use' counts as academic misconduct might vary on an individual basis, a potential remedy involves implementing barriers that make the pre-processing described in Table \ref{changes} more demanding than the assignments themselves. For example, incorporating plots into the Jupyter notebooks for data interpretation, rather than using tables, could compel the AI to analyze visual data - which it may not do accurately - or necessitate a textual plot description, a process found to adversely GPT-4's performance \cite{fengAI}.

Contrary to earlier studies examining physics essays \cite{yeadon2024evaluating}, which found that human evaluators could not distinguish AI-generated content from human work better than random chance, our findings reveal that for coding assignments markers can quite successfully identify AI work. This difference is not merely due to the lower scores of GPT-3.5 inputs; as detailed in Section \ref{sec-id}, all submissions with AI contributions were predominantly classified as either 'Definitely AI' or 'Probably AI'. Evaluators particularly noted the AI-generated plots' tendency to appear slightly askew or misaligned, attributes easily spotted by humans, such as unusual font size choices and or positioning. This said, a distinctive feature of student work highlighted by markers was the unique, sometimes bold, design choices students made, including unconventional color schemes in their plots. This contrasted sharply with the AI's preference for standard tableau colors in matplotlib. These observations suggest that the nuanced, creative decisions by students serve as a clear differentiator from AI-generated content.

\subsection{Limitations}
An important methodological concern of this study is the effect of the pre-processing (see Table \ref{changes}) on the AI's output quality. Although we deliberately limited the pre-processing to essentials, it played a crucial role in preparing the AI to comprehend and execute the given tasks effectively. By employing prompt engineering techniques, we noted a marked improvement in the AI's performance that was statistically significant. Considering further refinement of prompt engineering, such as incorporating detailed, step-by-step instructions, might boost the LLM's effectiveness even further. However, as previously discussed, this approach shifts the assessment focus from the inherent abilities of the AI to the efficiency of human-augmented AI interaction. Given that prior studies (e.g. Ghassemi et al. \cite{mollickCentaur}) have shown how user involvement can significantly vary the performance enhancements achieved with AI, this introduces an additional layer of complexity to any analysis. 

\section{Conclusion}
This study found that the latest LLMs have not surpassed human proficiency in physics coding assignments. Nonetheless, we observed a strict superiority in the capabilities of GPT-4 over GPT-3.5, and identified that prompt engineering can significantly enhance performance. Should the improvement trajectory from GPT-3.5 to GPT-4 continue, LLMs may soon outperform student capabilities. Additionally, our analysis revealed that plots generated by LLMs are distinguishable from student-created ones due to their often misaligned or skewed layouts and a tendency to utilize default color schemes. In contrast, student plots occasionally feature more unique, albeit sometimes garish, design choices. This distinction underscores the potential for human markers to identify AI-generated content. Crucially, this research highlights the importance of accounting for the degree of human intervention in evaluating the effectiveness of any LLM. As we move forward, these findings prompt a reevaluation of how we measure AI performance and the role of human collaboration in harnessing AI's full potential.  

\printbibliography

\newpage
\appendix
\section{Breakdown of marks by marker}\label{byMarker}
Plotting the results in Figure \ref{fig-scores} but by the scores from each of the three markers reveals strong similarity among them. To further investigate potential differences, we conducted an analysis of variance (ANOVA) on the three averages. The ANOVA yielded an F-value of 2.72 and a p-value of 0.067, which exceeds the conventional significance threshold of 0.05. Consequently, we fail to reject the null hypothesis, leading us to conclude that the group means are statistically equivalent.

\begin{figure}[!htbp]
\centering
\includegraphics[width=15cm]{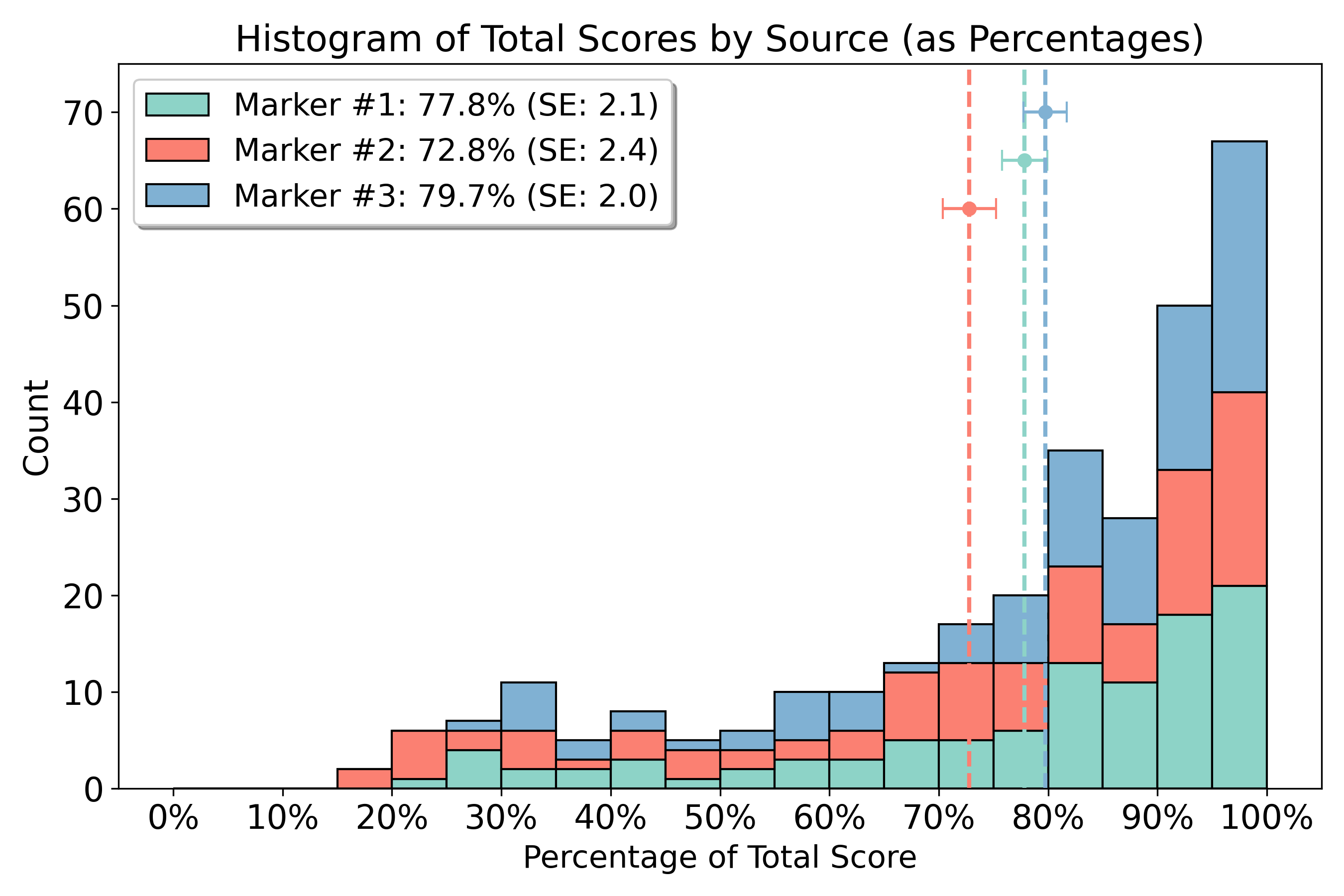}
\caption{Stacked histogram of the scores awarded by the three independent markers. Both the ANOVA and ICC models used find that the markers are consistent in their evaluations.}
\label{fig-marker-stacked}
\end{figure}

Beyond the ANOVA test, we employed the Intraclass Correlation Coefficient (ICC) analysis to evaluate the consistency of grading across multiple markers. The ICC1 model was used to assess the absolute agreement among markers and the ICC2 model was used to examine the agreement among markers on the relative ranking of submissions, rather than focusing on the exact scores assigned. The ICC ranges from -1 to 1, with a value of 1 indicating perfect agreement among markers, and -1 signifying complete disagreement. As shown in Table \ref{icc-table}, both ICC1 and ICC2 yielded a high value of 0.932, indicating an excellent agreement and consistency in the ratings provided by the markers. Furthermore, the high F values (41.942 for ICC1 and 68.293 for ICC2) and extremely low p-values ($\approx 0$) suggest that the observed variances are statistically significant, providing strong evidence for the reliability of the grading process.

\begin{table}[ht]
\centering
\caption{Intraclass Correlation Coefficient (ICC) Analysis Results}
\begin{tabular}{@{}lllllll@{}}
\toprule
Model & ICC & F & df1 & df2 & p-value & 95\% CI \\
\midrule
ICC1 & 0.932 & 41.942 & 99 & 200 & 0.0 & [0.91, 0.95] \\
ICC2 & 0.932 & 68.293 & 99 & 198 & 0.0 & [0.83, 0.97] \\
\bottomrule
\end{tabular}
\label{icc-table}
\end{table}

\end{document}